\documentclass[letterpaper]{article} % DO NOT CHANGE THIS

\ifdefined\aaaianonymous
    \usepackage[submission]{aaai2026}  % Anonymous submission version
\else
    \usepackage{aaai2026}              % Camera-ready version
\fi

\usepackage{times}  % DO NOT CHANGE THIS
\usepackage{helvet}  % DO NOT CHANGE THIS
\usepackage{courier}  % DO NOT CHANGE THIS
\usepackage[hyphens]{url}  % DO NOT CHANGE THIS
\usepackage{graphicx} % DO NOT CHANGE THIS
\usepackage{natbib}  % DO NOT CHANGE THIS AND DO NOT ADD ANY OPTIONS TO IT
\usepackage{caption} % DO NOT CHANGE THIS AND DO NOT ADD ANY OPTIONS TO IT
\usepackage{algorithm}
\usepackage{algorithmic}

\usepackage{newfloat}
\usepackage{listings}

\usepackage{pifont}
\usepackage{booktabs}
\usepackage{multirow} % 
\usepackage{colortbl} % 
\usepackage[table,xcdraw]{xcolor}   % 
\usepackage{makecell} 
\usepackage{bbding}
\usepackage{pifont}
\usepackage{tikz}
\usepackage{fancybox}
\usepackage{amsmath} 
\usepackage{amssymb}
\usepackage{makecell}

\DeclareCaptionStyle{ruled}{labelfont=normalfont,labelsep=colon,strut=off} % DO NOT CHANGE THIS
\floatstyle{ruled}
\newfloat{listing}{tb}{lst}{}
\floatname{listing}{Listing}

\title{ Exploring Modality-Aware Fusion and Decoupled Temporal Propagation for Multi-Modal Object Tracking }

\author {
    Shilei Wang\textsuperscript{\rm 1},
    Pujian Lai\textsuperscript{\rm 1},
    Dong Gao\textsuperscript{\rm 1},
    Jifeng Ning\textsuperscript{\rm 2},
    Gong Cheng\textsuperscript{\rm 1}\thanks{Corresponding author}
}
\affiliations {
    \textsuperscript{\rm 1}School of Automation, Northwestern Polytechnical University\\
    \textsuperscript{\rm 2}College of Information Engineering, Northwest A\&F University\\
    \{shileiwang, laipujian, 2019302284\}@mail.nwpu.edu.cn, njf@nwsuaf.edu.cn, gcheng@nwpu.edu.cn\\
    
}

\usepackage{bibentry}
\begin{document}

\maketitle

\begin{abstract}
Most existing multi-modal trackers adopt uniform fusion strategies, overlooking the inherent differences between modalities. Moreover, they propagate temporal information through mixed tokens, leading to entangled and less discriminative temporal representations.
To address these limitations, we propose MDTrack, a novel framework for modality-aware fusion and decoupled temporal propagation in multi-modal object tracking.
Specifically, for modality-aware fusion, we allocate dedicated experts to each modality (Infrared, Event, Depth, and RGB) to process their respective representations. The gating mechanism within the Mixture of Experts (MoE) then dynamically selects the optimal experts based on the input features, enabling adaptive and modality-specific fusion.
For decoupled temporal propagation, we introduce two separate State Space Model (SSM) structures to independently store and update the hidden states $h$ of the RGB and X-modal streams, effectively capturing their distinct temporal information. To ensure synergy between the two temporal representations, we incorporate a set of cross-attentions between the input features of the two SSMs, facilitating implicit information exchange. The resulting temporally enriched features are then integrated into the backbone via another set of cross-attention, enhancing MDTrack’s ability to leverage temporal information. 
Extensive experiments demonstrate the effectiveness of our proposed method. Both MDTrack-S (Modality-Specific Training) and MDTrack-U (Unified-Modality Training) achieve state-of-the-art performance across five multi-modal tracking benchmarks. The code is publicly available at https://github.com/wsumel/MDTrack
\end{abstract}

\section{Introduction}

\begin{figure}
    \centering
    \includegraphics[width=1.05\linewidth]{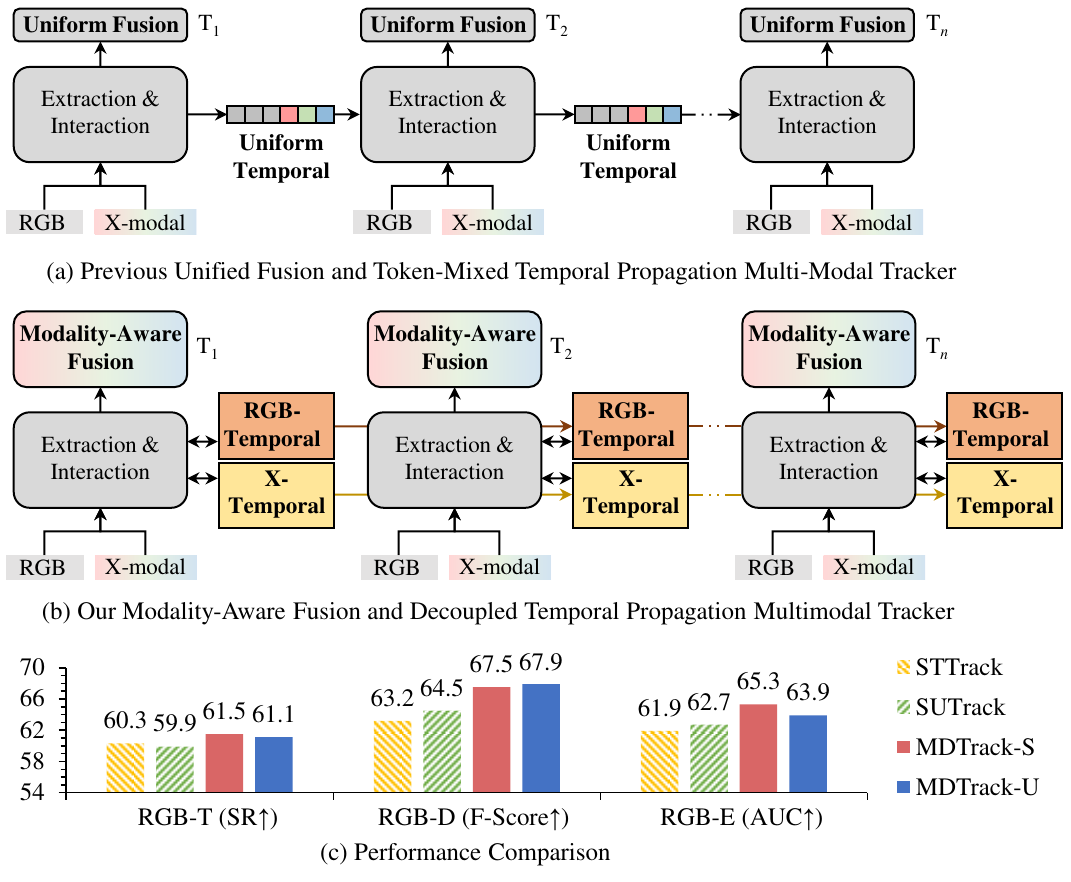}
    \caption{Overview of multi-modal tracking frameworks (a) and (b), with performance comparison (c) from left to right: STTrack, SUTrack, MDTrack-S, and MDTrack-U.}
    \label{fig:enter-intro}
\end{figure}

Visual object tracking (VOT) is a fundamental task in computer vision that aims to continuously localize an object in a video based on its initial position and has been widely used in autonomous driving, robotics, surveillance, and augmented reality. While recent RGB-based trackers~\cite{atom,DiMP,satrk,MCTrack,TransT,TrSiamTrDimp} have achieved impressive accuracy under normal conditions, they still struggle in challenging scenarios such as low illumination, motion blur, occlusion, and textureless backgrounds where appearance cues are unreliable. To address these limitations, multi-modal tracking has emerged as a promising paradigm by incorporating complementary sensor modalities such as infrared (IR), event, and depth data alongside RGB inputs~\cite{tbsi,bat,sdstrack,untrack}. For instance, IR effectively captures thermal signatures under poor lighting, event cameras detect rapid motion changes with high temporal resolution, and depth sensors provide geometric structure invariant to appearance variations. Consequently, multi-modal tracking offers a powerful solution to overcoming the inherent weaknesses of RGB-based methods and achieving reliable performance across diverse real-world environments.

Despite these advances, existing state-of-the-art multi-modal trackers~\cite{sttrack, sutrack}, as illustrated in Fig.~\ref{fig:enter-intro}(a), predominantly adopt a uniform fusion strategy that overlooks modality-specific differences. They use the same fusion module to integrate RGB+infrared (IR), RGB+event, or RGB+depth data, ignoring their distinct signal characteristics, noise patterns, and semantic properties. 
Under unified-modality training, this “one-size-fits-all” approach limits fusion adaptability and hinders the effective exploitation of each modality’s distinct strengths, ultimately resulting in suboptimal tracking performance.

Moreover, for temporal modeling, these trackers typically propagate temporal features through mixed tokens, following the paradigm of RGB-only trackers~\cite{zheng2024odtrack,MCTrack}. However, this entangles heterogeneous temporal dynamics, as RGB streams encode appearance and texture changes while X-modal streams (IR, event, depth) capture thermal stability, polarity events, or geometric consistency. Mixing these distinct temporal cues within a single propagation path causes mutual interference and confounded representations, ultimately impeding robust tracking under challenging scene variations.

%%%%%%%%%%%%%%%%%%%%%%%%%%%%%

To address the limitations of existing multi-modal trackers, we propose MDTrack, which integrates modality-aware fusion with decoupled temporal propagation, as illustrated in Fig.~\ref{fig:enter-intro}(b). This design fully exploits the unique characteristics of each modality while preserving their distinct temporal dynamics, enabling robust multi-modal object tracking.

Specifically, MDTrack adopts a Mixture of Experts (MoE)~\cite{MoE17,MoE21} framework for modality-aware fusion, where dedicated experts are assigned to IR, Event, Depth, and RGB modalities. A gating mechanism dynamically selects appropriate experts based on input features, enabling effective modality-specific fusion.
For decoupled temporal propagation, MDTrack employs two independent State Space Models (SSMs)~\cite{gu2023mamba} to maintain and update the hidden states of the RGB and X-modal streams separately, thereby modeling their distinct temporal dynamics without interference. In addition, cross-attention is applied to the input features to facilitate implicit inter-stream information exchange.
Together, these designs enhance inter-modal collaboration and temporal modeling, resulting in accurate and robust multi-modal tracking performance.

Performance comparisons on three multi-modal tracking tasks, as shown in Fig.~\ref{fig:enter-intro}(c), demonstrate that MDTrack significantly outperforms previous methods such as STTrack~\cite{sttrack} and SUTrack~\cite{sutrack}.

The main contributions of this work are summarized as follows:

\begin{itemize}
    \item We propose MDTrack, a novel multi-modal tracking paradigm that combines modality-aware fusion with decoupled temporal propagation to improve tracking robustness across diverse scenarios.
    \item We develop a modality-aware fusion based on an MoE, which dynamically selects dedicated experts for each modality to achieve effective cross-modal integration.
    \item We design a decoupled temporal propagation scheme that employs two independent SSMs for RGB and X-modal streams, allowing separate temporal dynamics modeling, while leveraging bidirectional cross-attention to achieve synchronized temporal reasoning and enriched temporal-contextual features.
    \item Extensive experiments on five mainstream multi-modal tracking benchmarks demonstrate that both MDTrack-S (Modality-Specific Training) and MDTrack-U (Unified-Modality Training) achieve state-of-the-art performance, validating the effectiveness and robustness of our proposed framework.
\end{itemize}

\section{Related Work}

\paragraph{Multi-Modal Object Tracking.} 
Recent advances in RGB-based visual tracking \cite{atom,ostrack,satrk,MCTrack,wang2025multi} have achieved remarkable performance with deep neural architectures. However, their robustness degrades in challenging scenarios such as low illumination, occlusion, and textureless regions, where appearance cues alone become unreliable.
To address these limitations, multi-modal tracking leverages complementary sensory modalities, including infrared, depth and event. TBSI~\cite{tbsi} enhances RGB-T tracking through temporal-bilateral semantic interactions, while DepthTrack~\cite{depthtrack} and ProTrack~\cite{protrack} improve RGB-D tracking by integrating geometric and visual cues. More recently, unified multi-modal frameworks have emerged as promising solutions: ViPT~\cite{vipt} employs prompt-based fusion but underutilizes non-RGB information, whereas SDSTrack~\cite{sdstrack}, BAT~\cite{bat}, and STTrack~\cite{sttrack} adopt symmetric architectures with temporal modeling to enhance robustness.

Despite these advances, existing methods predominantly adopt uniform fusion strategies and mixed-token temporal propagation, overlooking modality-specific differences and leading to entangled temporal representations. This highlights the need for more flexible and adaptive frameworks capable of achieving modality-aware fusion while decoupling temporal modeling, motivating the design of our proposed MDTrack.

\begin{figure*}
    \centering
    \includegraphics[width=1\linewidth]{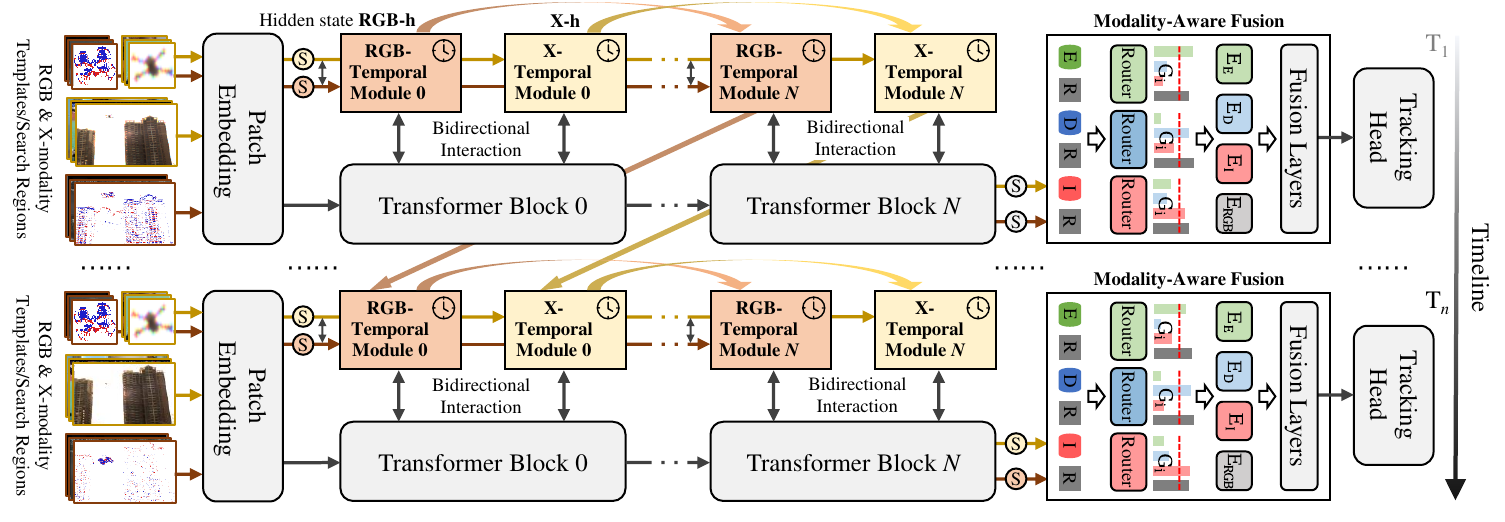}
    \caption{Overall tracking framework of MDTrack. Template and search region tokens from different modalities are concatenated and fed into the backbone. The search region tokens are then decoupled and stored separately for temporal propagation. Finally, the resulting features are fused by the modality-aware fusion module and passed to the tracking head for prediction.
    }
    \label{fig:enter-method}
\end{figure*}

\paragraph{Mixture of Experts.} 
The Mixture of Experts (MoE) method, introduced by Shazeer et al. ~\cite{MoE17}, dynamically selects specialized experts to handle tasks, significantly increasing model capacity without adding computational complexity. MoE’s core advantage is its ability to assign experts to different sub-tasks, improving efficiency while maintaining low cost. Lepikhin et al. ~\cite{lepikhin2020gshard} and Fedus et al. ~\cite{MoE21} extended MoE to Transformer architectures, enabling larger-scale pre-trained models.
MoE has been widely adopted in multi-modal learning. Ma et al. ~\cite{ma2018modeling} proposed a multi-gated MoE method for multi-task learning, assigning experts to specific tasks. Mustafa et al. ~\cite{mustafa2022multimodal} demonstrated MoE’s potential in vision-language model training, enhancing cross-modal alignment. These applications show that MoE excels not only in single-task optimization but also in multi-modal fusion tasks.
In multi-modal image fusion, Zhu et al.~\cite{zhu2024task} proposed TC-MoA, where expert modules serve as adapters to customize tasks without additional computational overhead. Building on this idea, MDTrack adopts an MoE-based fusion in which each modality is guided by a dedicated expert, enabling dynamic expert selection to improve fusion quality and tracking performance.

\paragraph{State Space Models.} The structured State Space Model (SSM) family, especially Mamba introduced by Gu et al. ~\cite{gu2023mamba} based on S4, has recently drawn attention in vision due to its ability to model long-range dependencies with linear complexity. Following this, Vision Mamba (ViM)~\cite{visionmamba} used bidirectional blocks to efficiently adapt SSM to visual data, achieving strong performance while significantly reducing memory usage compared to traditional ViT~\cite{vit}. VMamba~\cite{liu2024vmamba} further refined this approach by incorporating a 2D scanning SSM, enabling spatially aware sequence modeling with high efficiency.
In the single-RGB tracking domain, previous work MCITrack~\cite{mcitrack} employed Mamba SSMs within a HiViT backbone to model long-term sequence dependencies, enabling robust tracking over extended frames. This line of work validates the strong applicability of SSMs in capturing temporal context for visual tracking. Inspired by these advances, MDTrack integrates dual SSM modules to decouple and specialize temporal modeling for RGB and X-modal streams, benefiting from the temporal efficiency and expressiveness of Mamba-style dynamics.

\section{Method}

In this section, we first present an overview of the overall architecture of our proposed MDTrack. We then describe the decoupled temporal propagation module, which independently models the temporal dynamics of RGB and X-modal streams. Next, we introduce the modality-aware fusion strategy based on an MoE, which adaptively integrates cross-modal features. Finally, we detail the design of the tracking head.

\subsection{Overview}

The overall architecture of MDTrack is illustrated in Fig. \ref{fig:enter-method}. It comprises four key components: a backbone network for visual feature extraction, two decoupled temporal modules (RGB-temporal module and X-temporal module) for independent temporal propagation of RGB and X-modal streams, a modality-aware fusion module for modality-aware feature integration, and a tracking head for final target prediction.

MDTrack takes two video modalities as input, which jointly participate in the object tracking decision process. Specifically, for each modality, input frames are first transformed into template tokens ($\mathbf{Z}_{\text{RGB}}$, $\textbf{Z}_\text{X}$) and search tokens ($\mathbf{S}_{\text{RGB}}$, $\textbf{S}_\text{X}$) through patch embedding and positional encoding. These tokens are then concatenated along the spatial dimension and fed into the backbone to extract unified feature representations.

The backbone consists of $N$ stacked blocks, each paired with its corresponding temporal module. The temporal modules interact bidirectionally with the backbone: temporal features are injected into the backbone to enhance feature accuracy, while the backbone features are used to update the hidden states within the temporal modules, capturing distinct temporal dynamics of each modality.

Subsequently, the extracted features are refined and fused by the modality-aware fusion module, which identifies the modality types and adaptively integrates them using dedicated experts. Finally, the fused features are passed into the tracking head to predict the target location in the current frame.

\subsection{Decoupled Temporal Propagation}

At each time step $t$, we first process the RGB and X-modal templates $\mathbf{Z}_{\text{RGB}} \in \mathbb{R}^{N \times 3 \times H_\text{z} \times W_\text{z}}$ and $\mathbf{Z}_{\text{X}} \in \mathbb{R}^{N \times 3 \times H_\text{z} \times W_\text{z}}$, as well as the corresponding search regions $\mathbf{S}_{\text{RGB}} \in \mathbb{R}^{N \times 3 \times H_\text{s} \times W_\text{s}}$ and $\mathbf{S}_{\text{X}} \in \mathbb{R}^{N \times 3 \times H_\text{s} \times W_\text{s}}$. Each input is divided into patches and transformed into token sequences using the patch embedding strategy of HiViT, which progressively downsamples the inputs to better preserve spatial information.

The embedded tokens from both modalities are then concatenated along the spatial dimension and fed into the HiViT backbone, which extracts multi-scale features and models their contextual relationships through its four-stage architecture. To capture temporal dynamics in a modality-specific manner, the search tokens of the RGB and X-modal streams are propagated through their respective temporal modules at each stage.

Specifically, within each stage, the search tokens from both modalities first undergo a bidirectional cross-attention operation to enable implicit feature exchange while maintaining their modality-specific representations. These updated tokens are then fed into their respective Mamba layers based on SSMs to encode and update distinct temporal information for each modality.

The SSM is formulated as a linear dynamical system inspired by continuous-time SSMs, defined as:

\begin{equation}
\begin{aligned}
h' &= \textbf{A} h + \textbf{B} \textbf{S}_i, \\
\textbf{S}_i' &= \textbf{C} h + \textbf{D} \textbf{S}_i,
\end{aligned}
\quad i \in \{\text{RGB}, \text{X}\}.
\end{equation}
where $h$ denotes the hidden state, $\textbf{S}_i$ the input, and $\textbf{S}'_i$ the output, with $\textbf{A}$, $\textbf{B}$, $\textbf{C}$, and $\textbf{D}$ being learnable system parameters. To implement SSM in a deep network, it is discretized using the zero-order hold method:

\begin{equation}
\bar{\textbf{A}} = \exp(\Delta \textbf{A}), \quad \bar{\textbf{B}} = (\Delta \textbf{A})^{-1}(\exp(\Delta \textbf{A}) - \textbf{I}) \cdot \Delta \textbf{B} \approx \Delta \textbf{B},
\end{equation}
where $\Delta$ is the time scale parameter. The discretized SSM update equations are thus represented as:

\begin{equation}
\begin{aligned}
h^t &= \bar{\textbf{A}} h^{t-1} + \bar{\textbf{B}} \textbf{S}^t_i, \\ 
\textbf{S}'^t_i &= \textbf{C} h_t + \textbf{D} \textbf{S}^t_i,
\end{aligned}
\quad i \in \{\text{RGB}, \text{X}\}.
\end{equation}

Here, $h_{t-1}$ stores historical hidden states that carry crucial temporal context, $h_t$ is the updated hidden state based on the current input $\textbf{S}^t_i$, and $\textbf{S}'^t_i$ integrates temporal information for downstream tasks. In our framework, modality-specific SSMs independently update the hidden states $h_{\text{RGB}}^t$ and $h_\text{X}^t$, thereby preserving the unique temporal dynamics of each modality.

Subsequently, the search tokens carrying temporal information interact twice with the backbone features via cross-attention modules. This bidirectional interaction injects temporal context into the backbone representations to enhance feature accuracy, while also updating the search tokens with enriched spatial-semantic information.

At the next time step $t+1$, the updated hidden states $h_{\text{RGB}}^t$ and $h_\text{X}^t$ are propagated as $h_{\text{RGB}}^{t+1}$ and $h_\text{X}^{t+1}$, enabling MDTrack to efficiently and decoupledly transmit long-term temporal information across video frames. This decoupled temporal modeling design ensures that the RGB and X-modal streams maintain their unique temporal dynamics without interference, resulting in more robust multi-modal tracking performance.

\subsection{Modality-Aware Fusion Module}

As illustrated in Fig.~\ref{fig:enter-method}, the modality-aware fusion module consists of a modality expert library $\{\text{E}_{\text{RGB}}, \text{E}_{\text{T}}, \text{E}_{\text{E}}, \text{E}_{\text{D}}\}$, which contains modality-specific processing methods, a gating weight library $\{\textbf{G}_{\text{RGB}}, \textbf{G}_{\text{T}}, \textbf{G}_{\text{E}}, \textbf{G}_{\text{D}}\}$ to determine the contribution of each expert, and guided fusion weights $\textbf{F}_i$. This module operates in two main stages: modality-specific expert selection and expert-guided fusion.

In the modality-specific expert selection stage, we first concatenate the multi-modal features to form a unified representation of the RGB and X-modality token pairs, enabling cross-modal interactions in subsequent processing. The router then generates the gating weights $\textbf{G}_i$ for each modality expert based on the joint features. Specifically, the routing computation is defined as:

\begin{align}
\textbf{G}_i =  & \text{Softmax}(\text{TopK}(\textbf{S}_{\text{RGBX}} \cdot \textbf{W}_g  + N(0,1) \nonumber \\
 & \cdot \text{Softplus}(\textbf{S}_{\text{RGBX}} \cdot \textbf{W}_{\text{noise}}))),
\end{align}
where $\text{TopK}(\cdot)$ retains only the top $K$ ($K=2$) values, setting others to $-\infty$ so that their Softmax outputs become zero. $\textbf{W}_\text{g}$ and $\textbf{W}_{\text{noise}}$ are learnable parameters.

In the expert-guided fusion stage, each modality expert $\text{E}_i$ and its associated routing output $\textbf{G}_i$ are used to generate modality-aware fusion weights:

\begin{align}
\textbf{F}_{\mathrm{RGB}} &= \text{GAP}(\text{Sigmoid}(\textbf{G}_{\mathrm{RGB}} \cdot \text{E}_{\mathrm{RGB}}(\textbf{S}_{\mathrm{RGB}}))), \\
\textbf{F}_{\mathrm{X}} &= \text{GAP}(\text{Sigmoid}(\textbf{G}_{\mathrm{X}} \cdot \text{E}_{\mathrm{X}}(\textbf{S}_{\mathrm{X}}))),
\end{align}
where $\text{GAP}(\cdot)$ denotes global average pooling, and $\textbf{G}_i$ is the routing value for the corresponding adapter. The resulting modality-customized weights $\textbf{F}_i \in \mathbb{R}^{H \times W \times 1}$ range within $(0,1)$ and serve as an importance assessment for each modality's information.
We employ a load-balancing loss to regularize the routing process, thereby encouraging proper activation of experts. 

Finally, these weights refine and fuse the modality features via element-wise multiplication and weighted summation:

\begin{equation}
\textbf{S} = \lambda_{\text{RGB}} (\textbf{F}_{\text{RGB}} \odot \textbf{S}_{\text{RGB}}) + \lambda_{\text{X}} (\textbf{F}_{\text{X}} \odot \textbf{S}_{\text{X}}),
\end{equation}
where both $\lambda_{\text{RGB}}$ and $\lambda_{\text{X}}$ are set to 0.5. This operation effectively removes redundant information while preserving complementary features, producing an adaptive and modality-aware fused representation for tracking.

\begin{table*}[!t]
	\centering
    \fontsize{9pt}{10.5pt}\selectfont  % 设置字体为 9pt roman
    \setlength{\tabcolsep}{0.7mm}
    \renewcommand{\arraystretch}{1.2}
    
	% \resizebox{1\textwidth}{!}{
		\begin{tabular}{c |c |c |c c |c c| c c c |c c c}
			\toprule
			 & & & \multicolumn{2}{c|}{LasHeR} & \multicolumn{2}{c|}{RGBT234} & \multicolumn{3}{c|}{DepthTrack} & \multicolumn{3}{c}{VOT-RGBD2022} \\
			\cline{4-13}
			\multirow{15}{*}{\rotatebox{90}{Modality-Specific Training}} & \multicolumn{1}{c|}{\multirow{-2}{*}{Method}}&\multirow{-2}{*}{Publication}  & Pr ($\uparrow$) & AUC ($\uparrow$) & MPR ($\uparrow$) & MSR ($\uparrow$) & Pr ($\uparrow$) & Re ($\uparrow$) & F-score ($\uparrow$) & EAO ($\uparrow$) & Acc. ($\uparrow$) & Rob. ($\uparrow$) \\
			\midrule\midrule
			% &SGT~\cite{sgt} & MM'17 & 32.7 & 23.2 & 72.0 & 47.2 & -- & -- & -- & -- & -- & -- \\
			% &ATOM~\cite{atom} & CVPR'19 & - & - & - & - & - & - & - & 50.5 & 69.8 & 68.8 \\
			% &DiMP~\cite{DiMP} & ICCV'19 & - & - & - & - & - & - & - & 54.3 & 70.3 & 73.1 \\
			% &mfDiMP~\cite{mfdimp} & ICCVW'19 & 44.7 & 34.3 & 64.6 & 42.8 & - & - & - & - & - & - \\
			% &DAFNet~\cite{dafnet} & ICCVW'19 & 44.8 & 31.1 & 79.6 & 54.4 & - & - & - & - & - & - \\
			% DAL~\cite{dal} & ICPR 2021 & - & - & - & - & 51.2 & 36.9 & 42.9 & - & - & - \\
			% &DeT~\cite{depthtrack} & ICCV'21 & - & - & - & - & 56.0 & 50.6 & 53.2 & - & - & - \\

			% &DDiMP~\cite{vot2022} & ECCV'22 & - & - & - & - & 50.3 & 46.9 & 48.5 & - & - & - \\
			% &ProTrack~\cite{protrack} & MM'22 & 53.8 & 42.0 & 79.5 & 59.9 & 58.3 & 57.3 & 57.8 & 65.1 & 80.1 & 80.2 \\
			% &Ostrack~\cite{ostrack} & ECCV'22 & 51.5 & 41.2 & 72.9 & 54.9 & 53.6 & 52.2 & 52.9 & 67.6 & 80.3 & 83.3 \\
			&ViPT~\cite{vipt} & CVPR'23 & 65.1 & 52.5 & 83.5 & 61.7 & 59.2 & 59.6 & 59.4 & 72.1 & 81.5 & 87.1 \\
			&SPT~\cite{rgbd1k} & AAAI'23 & - & - & - & - & 52.7 & 54.9 & 53.8 & 65.1 & 79.8 & 85.1 \\
			
		    &Un-Track~\cite{untrack} & CVPR'24 & 66.7 & 53.6 & 83.7 & 61.8 & 61.3 & 61.0 & 61.2 & 72.1 & 81.5 & 87.1 \\

            &SDSTrack~\cite{sdstrack} & CVPR'24 & 66.5 & 53.1 & 84.8 & 62.5 & 61.9 & 60.9 & 61.4 & 72.8 & 81.2 & 88.3 \\

            &OneTracker~\cite{onetracker} & CVPR'24 & 67.2 & 53.8 & 85.7 & 64.2 & 60.7 & 60.4 & 60.9 & 72.7 & 81.9 & 87.2 \\
			
            &TBSI~\cite{tbsi} & CVPR'24 & 70.5 & 56.3 & 86.4 & 64.3 & - & - & - & - & - & - \\
            &TATrack~\cite{tattrack} & AAAI'24 & 70.2 & 56.1 & 87.2 & 64.4 & - & - & - & - & - & - \\
            &BAT~\cite{bat} & AAAI'24 & 70.2 & 56.3 &  86.8 & 64.1 & - & - & - & - & - & - \\
            &GMMT~\cite{GMMT} & AAAI'24 & 70.7 & 56.6 &  87.9 & 64.7 & - & - & - & - & - & - \\
            &STTrack~\cite{sttrack} &AAAI'25 & 76.0 & 60.3 & 89.8 & 66.7 &  63.3 & 63.4 & 63.2 &  77.6 & 82.5 & 93.7 \\

			% &MDTrack-S  & Ours & \textcolor{red}{76.5} & \textcolor{red}{61.4} & \textcolor{red}{93.0} & \textcolor{blue}{70.5} & \textcolor{blue}{67.5} & \textcolor{blue}{67.5} & \textcolor{blue}{67.5} & \textcolor{blue}{79.7} & \textcolor{blue}{83.6} & \textcolor{blue}{94.8} \\
            &MDTrack-S  & Ours & \textbf{\underline{76.5}} & \textbf{\underline{61.4}} & \textbf{\underline{93.0}} & \textbf{70.5} & \textbf{67.5} & \textbf{67.5} & \textbf{67.5} & \textbf{79.7} & \textbf{83.6} & \textbf{94.8} \\

            \cline{1-13}
            
            \multirow{9}{*}{\rotatebox{90}{Unified-Modality Training}} 
            &Stark~\cite{satrk} & ICCV‘21 & 41.8 & 33.3 & 67.7 & 49.6 & 39.7 & 40.6  & 38.8 & 44.5 & 71.4 & 59.8 \\
            &AiATrack~\cite{gao2022aiatrack} & ECCV'22 & 46.3 & 36.5 & 71.1 & 50.8 & 51.5 & 52.6  & 50.5 & 64.1 & 76.9 & 83.2 \\
            &OSTrack~\cite{ostrack} & ECCV'22 & 53.0 & 42.2 & 75.5 & 56.9 & 56.9 & 58.2 & 55.7 & 66.6 & 80.8 & 81.4 \\
            &SeqTrack~\cite{seqtrack} & CVPR'23 & 58.2 & 44.1 & 80.6 & 59.9 & 59.0 & 60.0 &  58.0  & 67.9 & 80.2 & 84.6 \\
            &ViPT~\cite{vipt} & CVPR'23 &60.8 & 49.0 & - & - & 56.1 & 56.2 & 56.0   & - & - & - \\
            
            &Un-Track~\cite{untrack} & CVPR'24 & 64.6 & 51.3 & 84.2 & 62.5 & 61.0 & 61.0 & 61.0 & 71.8 & 82.0 & 86.4 \\
            
            &SUTrack ~\cite{sutrack} &AAAI'25 & 74.5 & 59.9 & 92.2 & 69.5 &   65.1 & 65.7 & 64.5 &  76.5 & 82.8 & 91.8 \\

            &XTrack~\cite{tan2024xtrack} & ICCV'25 & 73.1 & 58.7 & 87.8 &65.4 & 65.4 & 64.3 &64.8 &74.0 & 82.8 & 88.9 \\
            % &MDTrack-U  & Ours & \textcolor{blue}{76.3 } & \textcolor{blue}{61.1} & \textcolor{blue}{92.6} & \textcolor{red}{70.6} & \textcolor{red}{68.1} & \textcolor{red}{67.6} & \textcolor{red}{67.9} & \textcolor{red}{80.0} & \textcolor{red}{83.5} & \textcolor{red}{95.1} \\
            &MDTrack-U  & Ours & \textbf{76.3} & \textbf{61.1} & \textbf{92.6} & \textbf{\underline{70.6}} & \textbf{\underline{68.1}} & \textbf{\underline{67.6}} & \textbf{\underline{67.9}} & \textbf{\underline{80.0}} & \textbf{\underline{83.5}} & \textbf{\underline{95.1}} \\

			\bottomrule
		\end{tabular}
	% }
    \caption{Comparisons with state-of-the-arts on LasHeR, RGBT234, DepthTrack, and VOT-RGBD2022. The best results are highlighted with bold underlines, and the second-best results are shown in bold fonts.}
	\label{table:1}
	
\end{table*}
\subsection{Head and Loss Function}

We adopt a prediction head architecture commonly used in recent Transformer-based trackers, with necessary adaptations for our multi-modal framework. Specifically, our tracking head comprises three parallel convolutional sub-networks dedicated to different prediction tasks. The first branch outputs the classification confidence map $\mathbf{P}_\text{S} \in \mathbb{R}^{1 \times \frac{H}{16} \times \frac{W}{16}}$, indicating the likelihood of the target at each spatial location. The second branch predicts the target's width and height $\mathbf{P}_\text{B} \in \mathbb{R}^{2 \times \frac{H}{16} \times \frac{W}{16}}$, while the third branch estimates the bounding box offsets $\mathbf{P}_\text{O} \in \mathbb{R}^{2 \times \frac{H}{16} \times \frac{W}{16}}$ to refine localization precision.

% During training, we formulate the objective as a weighted combination of classification and regression losses. The classification branch is supervised with a focal loss to address class imbalance and emphasize hard examples. For the regression branches, we employ a hybrid loss comprising an $\ell_1$ loss for bounding box parameter regression and a generalized IoU (GIoU) loss to improve overlap consistency with ground truth. Formally, the total loss for a training batch is defined as:

% \begin{equation}
% \mathcal{L} = \lambda_{\text{cls}} \mathcal{L}_{\text{cls}} + \lambda_{\text{l1}} \mathcal{L}_{\text{l1}} + \lambda_{\text{giou}} \mathcal{L}_{\text{giou}},
% \end{equation}
% where $\mathcal{L}_{\text{cls}}$, $\mathcal{L}_{\text{l1}}$, and $\mathcal{L}_{\text{giou}}$ denote classification, $\ell_1$, and GIoU losses, respectively. The coefficients $\lambda_{\text{cls}}$, $\lambda_{\text{l1}}$, and $\lambda_{\text{giou}}$ are set to 1, 5, and 2 by default to balance the learning objectives. 
% This design enables the joint optimization of target classification and precise bounding box regression within the multi-modal temporal fusion framework of MDTrack.

During training, we formulate the objective as a weighted combination of classification, regression, and load-balancing losses. The classification branch is supervised with a focal loss to address class imbalance and emphasize hard examples. For the regression branches, we employ a hybrid loss comprising an $\ell_1$ loss for bounding box parameter regression and a generalized IoU (GIoU) loss to improve overlap consistency with ground truth. Additionally, a load-balancing loss $\mathcal{L}_{\text{balance}}$ is introduced to regularize the routing process and encourage proper expert activation. Formally, the total loss for a training batch is defined as:

\begin{equation}
\mathcal{L} = \lambda_{\text{cls}} \mathcal{L}_{\text{cls}} + \lambda_{\text{l1}} \mathcal{L}_{\text{l1}} + \lambda_{\text{giou}} \mathcal{L}_{\text{giou}} + \lambda_{\text{balance}} \mathcal{L}_{\text{balance}},
\end{equation}
where $\mathcal{L}{\text{cls}}$, $\mathcal{L}{\text{l1}}$, $\mathcal{L}{\text{giou}}$, and $\mathcal{L}{\text{balance}}$ denote the classification, $\ell_1$, GIoU, and load-balancing losses, respectively. The coefficients $\lambda_{\text{cls}}$, $\lambda_{\text{l1}}$, $\lambda_{\text{giou}}$, and $\lambda_{\text{balance}}$ weight the contributions of the corresponding objectives during optimization.
% This design enables the joint optimization of target classification, precise bounding box regression, and balanced expert utilization within the multi-modal temporal fusion framework of MDTrack.

\section{Experiments}

%在本节中，我们首先详细阐述所提出的MDTrack模型的实现细节包括训练流程以及推理过程。随后，我们在多个模态跟踪测试基准上将MDTrack与其他领先方法进行对比分析。最后我们对MDTrack中的关键组件以及相应的超参数进行了消融分析。

In this section, we first present the implementation details of the proposed MDTrack model, including its training pipeline and inference strategy. We then conduct comprehensive comparisons with state-of-the-art methods on various multi-modal tracking benchmarks to evaluate its effectiveness. Finally, we perform ablation studies on the key components of MDTrack to analyze their contributions to the overall tracking performance.

\subsection{Implementation Details}

\textbf{Training.} We adopt two training strategies for MDTrack. The first follows the conventional multi-modal tracking setup, where the model is trained using only one modality at a time (i.e., RGB+Depth, RGB+Thermal, or RGB+Event). The second strategy merges all modality datasets for joint training, enabling a single model to handle tracking tasks across all modalities. We initialize parts of our model parameters using the pretrained weights from the RGB tracker \cite{mcitrack}. For all training settings, the template size is set to 112 $\times$ 112 and the search region size to 224 $\times$ 224. Training is performed on four NVIDIA RTX 4090 GPUs, with modality-specific training running for 20 epochs and mixed-modality training running for 30 epochs. Each epoch consists of 60,000 sample pairs, with a batch size of 16. The AdamW optimizer is used with a learning rate of 5e-5.
%我们采用了两种方式来对MDTrack进行训练，一种是遵循传统的多模态跟踪器设置每次只利用一种模态来进行训练（RGB+深度、RGB+热成像或RGB+事件）。第二种是将所有模态的数据集合并进行训练，用一个模型解决所有模态的跟踪问题。在所有的训练模式中模版被设置为112x112，搜索区域被设置为224x224.训练过程在四块RTX 4090Ti上进行。模态特定模式训练20个epochs，模态混合训练30个epochs。每个训练周期包含60000个样本对，batch size大小为24.采用AdamW优化器，学习率设置为5e-5。我们遵循大多数多模态跟踪器的方法将RGB跟踪器xxx作为基线模型
\begin{table*}[htbp]
    \centering
    \fontsize{9pt}{10.5pt}\selectfont  % 设置字体为 9pt roman
    \setlength{\tabcolsep}{0.9mm}
    \renewcommand{\arraystretch}{1.2}
    % \renewcommand{\arraystretch}{1.2}
    % \resizebox{\textwidth}{!}{
    \begin{tabular}{l|cccccc|ccccccc}
        \toprule
        \multirow{2}{*}{Method} & \multicolumn{6}{c|}{Modality-Specific Train} & \multicolumn{6}{c}{Unified-Modality Train} \\
        \cline{2-13}
           &  ViPT & Un-Track & SDSTrack & OneTrack & STTrack & MDTrack-S
          & SeqTrack & ViPT & Un-Track & SUTrack & XTrack& MDTrack-U \\
        \midrule
        Precision ($\uparrow$) 
         &  75.8 & 76.3 & 76.7 & 76.7 & 78.6 & \underline{\textbf{82.2}}
        & 66.5 & 74.0 & 75.5 &  79.9 & 80.5 &\textbf{81.3}\\
        Success ($\uparrow$)
         &  59.2 & 59.7 & 59.7 & 60.8 & 61.9 & \underline{\textbf{65.3}}
        & 50.4 & 57.9 & 58.9 &62.7 & 63.3 &\textbf{63.9} \\
        \bottomrule
    \end{tabular}
    % }
    \caption{Precision and Success comparisons between Modality-Specific and Unified-Modality training on the VisEvent dataset. The best results are highlighted with bold underlines, and the second-best results are shown in bold fonts.}
    \label{tab:event_specific_vs_unimodel}
\end{table*}

\textbf{Inference.} During inference, the template and search region sizes remain consistent with those used during training. Temporal information is incrementally integrated into the tracking pipeline in a modality-decoupled manner. MDTrack-S and MDTrack-U achieve inference speed of approximately 25 frames per second (FPS) when tested on an NVIDIA RTX 4090 GPU.

% \begin{table*}[htbp]
%     \centering
%     \fontsize{9pt}{10.5pt}\selectfont  % 设置字体为 9pt roman
%     \setlength{\tabcolsep}{0.1mm}
%     \renewcommand{\arraystretch}{1.2}
%     % \renewcommand{\arraystretch}{1.2}
%     % \resizebox{\textwidth}{!}{
%     \begin{tabular}{l|ccccccc|cccccccc}
%         \toprule
%         \multirow{2}{*}{Method} & \multicolumn{7}{c|}{Modality-Specific Train} & \multicolumn{7}{c}{Unified-Modality Train} \\
%         \cline{2-15}
%           & TransT E &  ViPT & Un-Track & SDSTrack & OneTrack & STTrack & MDTrack-S
%           & OSTrack & SeqTrack & ViPT & Un-Track & SUTrack & XTrack& MDTrack-U \\
%         \midrule
%         Precision ($\uparrow$) 
%          & 65.0 &  75.8 & 76.3 & 76.7 & 76.7 & 78.6 & \underline{\textbf{82.2}}
%          & 69.1 & 66.5 & 74.0 & 75.5 &  79.9 & 80.5 &\textbf{81.3}\\
%         Success ($\uparrow$)
%         & 47.4 &  59.2 & 59.7 & 59.7 & 60.8 & 61.9 & \underline{\textbf{65.3}}
%          & 52.5 & 50.4 & 57.9 & 58.9 &62.7 & 63.3 &\textbf{63.9} \\
%         \bottomrule
%     \end{tabular}
%     % }
%     \caption{Precision and Success comparisons between Modality-Specific and Unified-Modality training on the VisEvent dataset. The best results are highlighted with bold underlines, and the second-best results are shown in bold fonts.}
%     \label{tab:event_specific_vs_unimodel}
% \end{table*}

\subsection{Comparison with State-of-the-Arts}

We conducted a comprehensive comparison of MDTrack against recent state-of-the-art multi-modal trackers across five benchmark datasets. All experimental results are obtained by running MDTrack once on each dataset, following the standard evaluation approach for most tracking methods. As shown in Tab. \ref{table:1} and Tab. \ref{tab:event_specific_vs_unimodel}, both MDTrack-S (the Modality-Specific Training variant) and MDTrack-U (the Unified-Modality Training variant) consistently achieved either the best or second-best results on all five datasets. Unlike Un-Track, which exhibits a noticeable performance gap between its two training modes, MDTrack-S and MDTrack-U deliver similarly strong performance, highlighting the effectiveness of the proposed modality-aware fusion design.

\textbf{LasHeR.} LasHeR is a large-scale RGBT tracking benchmark with 1,224 RGB-thermal video pairs and over 730K annotated frames~\cite{lasher}.  
On LasHeR, both MDTrack-S and MDTrack-U achieve state-of-the-art performance among their respective categories.  
MDTrack-S obtains 76.5\% precision and 61.4\% AUC, while MDTrack-U achieves 76.3\% precision and 61.1\% AUC, significantly outperforming all previous methods and setting new benchmarks in RGBT tracking.

\textbf{RGBT234.} RGBT234 contains 234 spatially aligned RGB-thermal video pairs with 234K annotated frames under diverse real-world conditions~\cite{rgbt234}.  
MDTrack-S delivers the best performance with an MPR of 93.0\% and MSR of 70.5\%, surpassing the previous best method STTrack by 3.2\% and 3.8\%, respectively.  
MDTrack-U also achieves strong results, reaching 92.6\% MPR and 70.6\% MSR, outperforming SUTrack (92.2\% / 69.5\%) and demonstrating excellent generalization under unified training.

\textbf{DepthTrack.} DepthTrack is the largest RGB-D tracking benchmark, consisting of 200 sequences across 90+ object classes and 40+ scenes~\cite{depthtrack}.  
MDTrack-S achieves balanced and robust performance with 67.5\% precision, 67.5\% recall, and 67.5\% F1-score, improving upon the previous best (STTrack, 63.2\% F1) by 4.3\%.  
MDTrack-U further advances the performance, attaining 68.1\% precision, 67.6\% recall, and 67.9\% F1-score, thereby setting new state-of-the-art results across all metrics on this dataset.

\textbf{VOT-RGBD2022.} VOT-RGBD2022 is a short-term RGB-D tracking benchmark featuring over 140 sequences and evaluated with EAO, accuracy, and robustness~\cite{vot2022}.  
MDTrack-S achieves 79.7\% EAO, 83.6\% accuracy, and 94.8\% robustness, ranking second among all methods.  
In contrast, MDTrack-U attains the highest EAO score of 80.0\%, with 83.5\% accuracy and 95.1\% robustness, surpassing all existing trackers and demonstrating superior performance under unified training.

\textbf{VisEvent.} VisEvent is the first large-scale benchmark for RGB-event tracking, with 820 synchronized video pairs and 371K annotated frames~\cite{visevent}.  
MDTrack-S achieves the highest Precision and Success of 82.2\% and 65.3\%, outperforming the previous best method STTrack by 3.6\% in Precision and 3.4\% in Success.  
Meanwhile, MDTrack-U obtains competitive performance with 81.3\% Precision and 63.9\% Success, ranking second overall and consistently outperforming other leading methods such as OneTrack, SUTrack, and Un-Track.

\subsection{Ablation Studies}

We conduct ablation studies on key components of MDTrack, including the modality-aware fusion module and the decoupled temporal propagation module. 
% These experiments provide an in-depth analysis of how each design choice contributes to the overall tracking performance.

\begin{table*}[htbp]
    \centering
    % \fontsize{9pt}{10.5pt}\selectfont  % 设置字体为 9pt roman
    \small
    \setlength{\tabcolsep}{3.5mm}
    \renewcommand{\arraystretch}{1.2}
    % \resizebox{\linewidth}{!}{
    \begin{tabular}{l|cccc}
        \toprule
        Configuration & LasHeR & DepthTrack & VisEvent & Mean \\
        \midrule
        Baseline & 58.5& 65.8& 62.2& 62.2 \\
        + Token-based Temporal  Module & 59.4 (+0.9) & 66.3 (+0.5) & 62.3 (+0.1) & 62.7 (+0.5) \\
        + Temporal Module & 59.3 (+0.8) & 66.6 (+0.8) & 62.6 (+0.4) & 62.8 (+0.6) \\
        + Decoupled Temporal Module & 60.2 (+1.7) & 67.6 (+1.8) & 63.3 (+1.1) & 63.7 (+1.5) \\
        + Fusion Module & 58.8 (+0.3) & 65.9 (+0.1) & 62.4 (+0.2) & 62.4 (+0.2) \\
        + Modality-Aware Fusion Module & 59.6 (+1.1) & 66.3 (+0.5) & 62.8 (+0.6) & 62.9 (+0.7) \\
        + Decoupled Temporal Module  \& Modality-Aware  Fusion Module & \textbf{61.1 (+2.6)} & \textbf{67.9 (+2.1)} & \textbf{63.9 (+1.7)} & \textbf{64.3 (+2.1)} \\
        \bottomrule
    \end{tabular}
    % }
    \caption{Ablation study of MDTrack on the LasHeR, DepthTrack, and VisEvent datasets. Each row illustrates a different module added to the baseline.}
    \label{tab:ablation}
\end{table*}

\textbf{Module Contribution Analysis.}  
From Tab.~\ref{tab:ablation}, adding the decoupled temporal module improves performance to 60.2\% (+1.7\%) AUC on LasHeR, 67.6\% (+1.8\%) F-score on DepthTrack, and 63.3\% (+1.1\%) Success rate on VisEvent, achieving a mean gain of +1.5\%. This demonstrates that decoupling temporal modeling effectively enhances target representation across diverse modalities.  
Incorporating the modality-aware fusion module achieves 59.6\% (+1.1\%) AUC on LasHeR, 66.3\% (+0.5\%) F-score on DepthTrack, and 62.8\% (+0.6\%) Success rate on VisEvent, with an average improvement of +0.7\%. This validates the advantage of expert-based adaptive fusion in leveraging complementary cross-modal cues.  
Finally, combining the decoupled temporal module and modality-aware fusion module achieves the best results, reaching 61.1\% (+2.6\%) AUC on LasHeR, 67.9\% (+2.1\%) F-score on DepthTrack, and 63.9\% (+1.7\%) Success rate on VisEvent, leading to a mean gain of +2.1\%. These results confirm that temporal decoupling and modality-aware fusion are complementary, jointly contributing to robust multi-modal tracking performance.

\textbf{Temporal Propagation Analysis.} 
We investigate three designs for temporal information propagation in MDTrack. As shown in Tab. \ref{tab:ablation}, using the Token-based Temporal Module, which concatenates historical template and search region tokens at the beginning of the video and propagates them directly across frames, yields only marginal improvements (e.g., +0.5\% on average). This limited gain is mainly due to the entanglement of temporal information between RGB and X modalities, leading to feature interference.
Similarly, introducing the Temporal Module based on a single SSM~\cite{gu2023mamba} for mixed temporal propagation brings slightly higher improvements (+0.6\% on average). However, the mixed design still cannot fully disentangle modality-specific temporal cues.
In contrast, the decoupled temporal module, which employs separate SSMs for RGB and X modalities along with implicit cross-attention between them, achieves the most significant performance boost (+1.5\% on average). This demonstrates that decoupling modality-specific temporal representations while enabling their interaction effectively enhances temporal modeling and overall tracking accuracy.

\textbf{Modality-aware Fusion Analysis.} 
We evaluate two fusion designs in MDTrack. As shown in Tab. \ref{tab:ablation}, employing the Fusion Module, which directly combines RGB and X modality features after processing them with a unified expert structure, brings only limited improvements (+0.2\% on average). This is because treating all modalities uniformly without considering their modality-specific differences leads to suboptimal fusion, as modality-specific cues may interfere with each other.
In contrast, the modality-aware fusion Module leverages an MoE mechanism to adaptively select different experts and guidance strategies based on the specific modality task. This design yields a higher performance gain (+0.7\% on average), demonstrating that adaptive expert selection effectively captures complementary information across modalities while suppressing irrelevant features. Overall, modality-aware fusion significantly enhances the robustness and generalization ability of our tracker in diverse multimodal scenarios.

\begin{figure}
    \centering
    \includegraphics[width=1\linewidth]{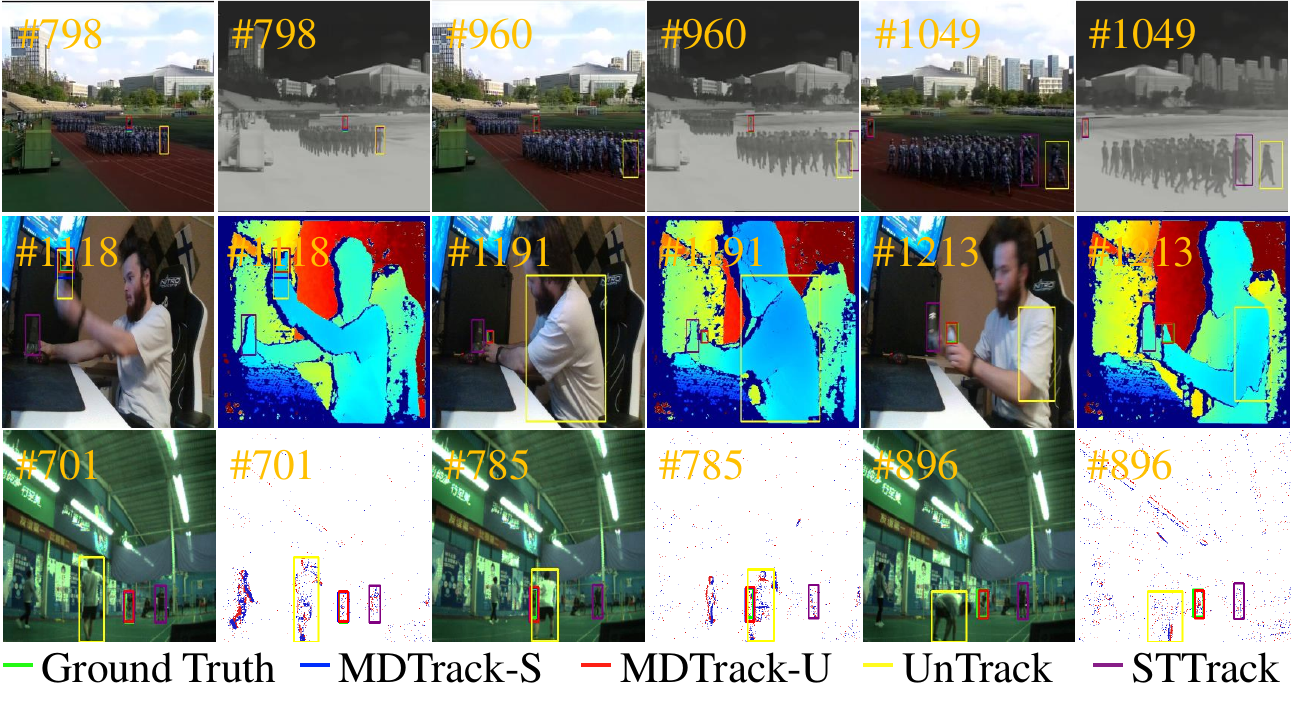}
    \caption{Visual comparisons of MDTrack-S and MDTrack-U with other multimodal trackers on the LasHeR, DepthTrack, and VisEvent datasets.}
    \label{fig:visual}
\end{figure}

\textbf{Visual Comparison.} 
As shown in Fig. \ref{fig:visual}, in the RGBT tracking scenario with numerous similar objects, MDTrack-S and MDTrack-U achieve stable tracking by effectively utilizing decoupled temporal propagation and modality-aware adaptive fusion, enabling them to accurately distinguish targets from distractors by leveraging both temporal cues and infrared information. In the RGBD tracking scenario, where the cup is partially occluded, MDTrack-S and MDTrack-U robustly localize the target by integrating depth information through their modality-specific experts while preserving temporal consistency via the decoupled temporal module. In the RGBE tracking scenario, characterized by fast-moving basketball players under dim lighting, MDTrack-S and MDTrack-U maintain accurate tracking by exploiting the high temporal resolution of event data alongside robust temporal modeling. These results demonstrate that the proposed decoupled temporal modeling and modality-aware fusion effectively improve tracking robustness and accuracy.

\section{Conclusion}

We present MDTrack, a novel multi-modal tracking framework that effectively addresses the challenges of modality heterogeneity and temporal entanglement. By integrating a modality-aware fusion mechanism based on an MoE with decoupled temporal propagation through dual structured SSMs, MDTrack captures modality-specific features and temporal dynamics while enabling their synergistic interaction. 
Extensive evaluations on five diverse benchmarks demonstrate that both modality-specific and unified training paradigms achieve state-of-the-art performance, underscoring the robustness and generalizability of our approach. MDTrack provides an effective approach for multi-modal tracking and offers valuable insights for future research on robust visual understanding using heterogeneous sensor data.

\section{Acknowledgments}
This work was supported in part by the National Natural Science Foundation of China under Grants 62376223 and 62476227, and in part by the Innovation Foundation for Doctor Dissertation of Northwestern Polytechnical University under Grant CX2025087, as well as the Fundamental Research Funds for the Central Universities.

% \end{acks}

\bibliography{aaai2026}

@String(CVPR= {IEEE Conf. Comput. Vis. Pattern Recog.})

@String(ICCV= {Int. Conf. Comput. Vis.})

@String(ECCV= {Eur. Conf. Comput. Vis.})

@String(TIP  = {IEEE Trans. Image Process.})

@String(ACMMM= {ACM Int. Conf. Multimedia})

@String(ICLR = {Int. Conf. Learn. Represent.})

@String(AAAI = {AAAI})

@String(TCYB = {IEEE Trans. Cybern.})

@String(CVPR  = {CVPR})

@String(ICCV  = {ICCV})

@String(ECCV  = {ECCV})

@String(TIP   = {IEEE TIP})

@String(ACMMM = {ACM MM})

@String(ICLR  = {ICLR})

@String(TCYB = {IEEE TCYB})

@inproceedings{satrk,
	author = {Yan, Bin and Peng, Houwen and Fu, Jianlong and Wang, Dong and Lu, Huchuan},
	title = {Learning spatio-temporal transformer for visual tracking},
	booktitle = ICCV,
	pages = {10448--10457},
	year={2021}
}

@inproceedings{wang2025multi,
  title={Multi-State Tracker: Enhancing Efficient Object Tracking via Multi-State Specialization and Interaction},
  author={Wang, Shilei and Cheng, Gong and Lai, Pujian and Gao, Dong and Han, Junwei},
  booktitle={Proceedings of the 33rd ACM International Conference on Multimedia},
  pages={4087--4096},
  year={2025}
}

@inproceedings{ostrack,
	title={Joint feature learning and relation modeling for tracking: A one-stream framework},
	author={Ye, Botao and Chang, Hong and Ma, Bingpeng and Shan, Shiguang and Chen, Xilin},
	booktitle=ECCV,
	pages={341--357},
	year={2022},
}

@inproceedings{ma2018modeling,
  title={Modeling task relationships in multi-task learning with multi-gate mixture-of-experts},
  author={Ma, Jiaqi and Zhao, Zhe and Yi, Xinyang and Chen, Jilin and Hong, Lichan and Chi, Ed H},
  booktitle={Proceedings of the 24th ACM SIGKDD international conference on knowledge discovery \& data mining},
  pages={1930--1939},
  year={2018}
}

@article{liu2024vmamba,
  title={Vmamba: Visual state space model},
  author={Liu, Yue and Tian, Yunjie and Zhao, Yuzhong and Yu, Hongtian and Xie, Lingxi and Wang, Yaowei and Ye, Qixiang and Jiao, Jianbin and Liu, Yunfan},
  journal={Advances in neural information processing systems},
  volume={37},
  pages={103031--103063},
  year={2024}
}

@article{visionmamba,
  title={Vision mamba: Efficient visual representation learning with bidirectional state space model},
  author={Zhu, Lianghui and Liao, Bencheng and Zhang, Qian and Wang, Xinlong and Liu, Wenyu and Wang, Xinggang},
  journal={arXiv preprint arXiv:2401.09417},
  year={2024}
}

@inproceedings{zhu2024task,
  title={Task-customized mixture of adapters for general image fusion},
  author={Zhu, Pengfei and Sun, Yang and Cao, Bing and Hu, Qinghua},
  booktitle={Proceedings of the IEEE/CVF conference on computer vision and pattern recognition},
  pages={7099--7108},
  year={2024}
}

@article{mustafa2022multimodal,
  title={Multimodal contrastive learning with limoe: the language-image mixture of experts},
  author={Mustafa, Basil and Riquelme, Carlos and Puigcerver, Joan and Jenatton, Rodolphe and Houlsby, Neil},
  journal={Advances in Neural Information Processing Systems},
  volume={35},
  pages={9564--9576},
  year={2022}
}

@inproceedings{atom,
	author = {Danelljan, Martin and Bhat, Goutam and Khan, Fahad Shahbaz and Felsberg, Michael},
	title = {Atom: Accurate tracking by overlap maximization},
	booktitle = CVPR,
	pages = {4660--4669},
	year={2019}
}

@inproceedings{zheng2024odtrack,
  title={Odtrack: Online dense temporal token learning for visual tracking},
  author={Zheng, Yaozong and Zhong, Bineng and Liang, Qihua and Mo, Zhiyi and Zhang, Shengping and Li, Xianxian},
  booktitle={Proceedings of the AAAI conference on artificial intelligence},
  volume={38},
  number={7},
  pages={7588--7596},
  year={2024}
}

@inproceedings{TransT,
  title={Transformer tracking},
  author={Chen, Xin and Yan, Bin and Zhu, Jiawen and Wang, Dong and Yang, Xiaoyun and Lu, Huchuan},
  booktitle={Proceedings of the IEEE/CVF Conference on Computer Vision and Pattern Recognition},
  pages={8126--8135},
  year={2021}
}

@inproceedings{TrSiamTrDimp,
  title={Transformer meets tracker: Exploiting temporal context for robust visual tracking},
  author={Wang, Ning and Zhou, Wengang and Wang, Jie and Li, Houqiang},
  booktitle={Proceedings of the IEEE/CVF Conference on Computer Vision and Pattern Recognition},
  pages={1571--1580},
  year={2021}
}

@inproceedings{DiMP,
	author = {Bhat, Goutam and Danelljan, Martin and Gool, Luc Van and Timofte, Radu},
	title = {Learning discriminative model prediction for tracking},
	booktitle = ICCV,
	pages = {6182--6191},
	year={2019}
}

@article{lepikhin2020gshard,
  title={Gshard: Scaling giant models with conditional computation and automatic sharding},
  author={Lepikhin, Dmitry and Lee, HyoukJoong and Xu, Yuanzhong and Chen, Dehao and Firat, Orhan and Huang, Yanping and Krikun, Maxim and Shazeer, Noam and Chen, Zhifeng},
  journal={arXiv preprint arXiv:2006.16668},
  year={2020}
}

@article{MoE17,
  title={Outrageously large neural networks: The sparsely-gated mixture-of-experts layer},
  author={Shazeer, Noam and Mirhoseini, Azalia and Maziarz, Krzysztof and Davis, Andy and Le, Quoc and Hinton, Geoffrey and Dean, Jeff},
  journal={arXiv preprint arXiv:1701.06538},
  year={2017}
}

@article{gu2023mamba,
  title={Mamba: Linear-time sequence modeling with selective state spaces},
  author={Gu, Albert and Dao, Tri},
  journal={arXiv preprint arXiv:2312.00752},
  year={2023}
}

@article{MoE21,
  title={Switch transformers: Scaling to trillion parameter models with simple and efficient sparsity},
  author={Fedus, William and Zoph, Barret and Shazeer, Noam},
  journal={Journal of Machine Learning Research},
  volume={23},
  number={120},
  pages={1--39},
  year={2022}
}

@inproceedings{onetracker,
	title={Onetracker: Unifying visual object tracking with foundation models and efficient tuning},
	author={Hong, Lingyi and Yan, Shilin and Zhang, Renrui and Li, Wanyun and Zhou, Xinyu and Guo, Pinxue and Jiang, Kaixun and Chen, Yiting and Li, Jinglun and Chen, Zhaoyu and others},
	booktitle=CVPR,
	pages={19079--19091},
	year={2024}
}

@inproceedings{seqtrack,
	title={Seqtrack: Sequence to sequence learning for visual object tracking},
	author={Chen, Xin and Peng, Houwen and Wang, Dong and Lu, Huchuan and Hu, Han},
	booktitle=CVPR,
	pages={14572--14581},
	year={2023}
}

@inproceedings{mcitrack,
  title={Exploring Enhanced Contextual Information for Video-Level Object Tracking}, 
  author={Ben Kang and Xin Chen and Simiao Lai and Yang Liu and Yi Liu and Dong Wang},
  booktitle={AAAI},
  year={2025}
}

@inproceedings{sttrack,
  title={Exploiting multimodal spatial-temporal patterns for video object tracking},
  author={Hu, Xiantao and Tai, Ying and Zhao, Xu and Zhao, Chen and Zhang, Zhenyu and Li, Jun and Zhong, Bineng and Yang, Jian},
  booktitle={Proceedings of the AAAI Conference on Artificial Intelligence},
  volume={39},
  number={4},
  pages={3581--3589},
  year={2025}
}

@misc{GMMT,
      title={Generative-based Fusion Mechanism for Multi-Modal Tracking}, 
      author={Zhangyong Tang and Tianyang Xu and Xuefeng Zhu and Xiao-Jun Wu and Josef Kittler},
      booktitle={AAAI Conference on Artificial Intelligence},
      year={2024} 
}

@article{tattrack, title={Target-Aware Tracking with Long-term Context Attention}, author={He, Kaijie and Zhang, Canlong and Xie, Sheng and Li, Zhixin and Wang, Zhiwen}, journal={arXiv preprint arXiv:2302.13840}, year={2023} }

@inproceedings{tbsi,
  title={Bridging Search Region Interaction With Template for RGB-T Tracking},
  author={Hui, Tianrui and Xun, Zizheng and Peng, Fengguang and Huang, Junshi and Wei, Xiaoming and Wei, Xiaolin and Dai, Jiao and Han, Jizhong and Liu, Si},
  booktitle={Proceedings of the IEEE/CVF Conference on Computer Vision and Pattern Recognition},
  pages={13630--13639},
  year={2023}
}

@inproceedings{depthtrack,
	title={Depthtrack: Unveiling the power of rgbd tracking},
	author={Yan, Song and Yang, Jinyu and K{\"a}pyl{\"a}, Jani and Zheng, Feng and Leonardis, Ale{\v{s}} and K{\"a}m{\"a}r{\"a}inen, Joni-Kristian},
	booktitle=ICCV,
	pages={10725--10733},
	year={2021}
}

@inproceedings{gao2022aiatrack,
  title={AiATrack: Attention in Attention for Transformer Visual Tracking},
  author={Gao, Shenyuan and Zhou, Chunluan and Ma, Chao and Wang, Xinggang and Yuan, Junsong},
  booktitle={European Conference on Computer Vision},
  pages={146--164},
  year={2022},
  organization={Springer}
}

@inproceedings{rgbd1k,
	title={RGBD1K: A large-scale dataset and benchmark for RGB-D object tracking},
	author={Zhu, Xue-Feng and Xu, Tianyang and Tang, Zhangyong and Wu, Zucheng and Liu, Haodong and Yang, Xiao and Wu, Xiao-Jun and Kittler, Josef},
	booktitle=AAAI,
	volume={37},
	number={3},
	pages={3870--3878},
	year={2023}
}

@inproceedings{sutrack,
  title={SUTrack: Towards Simple and Unified Single Object Tracking},
  author={Chen, Xin and Kang, Ben and Geng, Wanting and Zhu, Jiawen and Liu, Yi and Wang, Dong and Lu, Huchuan},
  booktitle=AAAI,
  year={2025}
}

@inproceedings{tan2024xtrack,
  title={XTrack: Multimodal Training Boosts RGB-X Video Object Trackers},
  author={Tan, Yuedong and Wu, Zongwei and Fu, Yuqian and Zhou, Zhuyun and Sun, Guolei and Ma, Chao and Paudel, Danda Pani and Van Gool, Luc and Timofte, Radu},
  booktitle = {Proceedings of the IEEE/CVF International Conference on Computer Vision (ICCV)},
  year={2025}
}

@inproceedings{vipt,
	title={Visual prompt multi-modal tracking},
	author={Zhu, Jiawen and Lai, Simiao and Chen, Xin and Wang, Dong and Lu, Huchuan},
	booktitle=CVPR,
	pages={9516--9526},
	year={2023}
}

@inproceedings{untrack,
	title={Single-model and any-modality for video object tracking},
	author={Wu, Zongwei and Zheng, Jilai and Ren, Xiangxuan and Vasluianu, Florin-Alexandru and Ma, Chao and Paudel, Danda Pani and Van Gool, Luc and Timofte, Radu},
	booktitle=CVPR,
	pages={19156--19166},
	year={2024}
}

@article{visevent,
	title={Visevent: Reliable object tracking via collaboration of frame and event flows},
	author={Wang, Xiao and Li, Jianing and Zhu, Lin and Zhang, Zhipeng and Chen, Zhe and Li, Xin and Wang, Yaowei and Tian, Yonghong and Wu, Feng},
	journal=TCYB,
	volume={54},
	number={3},
	pages={1997--2010},
	year={2023},
}

@inproceedings{sdstrack,
	title={Sdstrack: Self-distillation symmetric adapter learning for multi-modal visual object tracking},
	author={Hou, Xiaojun and Xing, Jiazheng and Qian, Yijie and Guo, Yaowei and Xin, Shuo and Chen, Junhao and Tang, Kai and Wang, Mengmeng and Jiang, Zhengkai and Liu, Liang and others},
	booktitle=CVPR,
	pages={26551--26561},
	year={2024}
}

@InProceedings{vit,
	author  = {Dosovitskiy, Alexey and Beyer, Lucas and Kolesnikov, Alexander and Weissenborn, Dirk and Zhai, Xiaohua and Unterthiner, Thomas and  Dehghani, Mostafa and Minderer, Matthias and Heigold, Georg and Gelly, Sylvain and Uszkoreit, Jakob and Houlsby, Neil},
	title  =   {An Image is Worth 16x16 Words: Transformers for Image Recognition at Scale},
	booktitle = ICLR,
	pages={1--22},
	year  = {2021}
}

@article{MCTrack,
  title={Modelling of Multiple Spatial-Temporal Relations for Robust Visual Object Tracking},
  author={Wang, Shilei and Wang, Zhenhua and Sun, Qianqian and Cheng, Gong and Ning, Jifeng},
  journal={IEEE Transactions on Image Processing},
  year={2024},
  publisher={IEEE}
}

@inproceedings{bat,
	title={Bi-directional adapter for multimodal tracking},
	author={Cao, Bing and Guo, Junliang and Zhu, Pengfei and Hu, Qinghua},
	booktitle=AAAI,
	volume={38},
	number={2},
	pages={927--935},
	year={2024}
}

@inproceedings{vot2022,
	title={The tenth visual object tracking vot2022 challenge results},
	author={Kristan, Matej and Leonardis, Ale{\v{s}} and Matas, Ji{\v{r}}{\'\i} and Felsberg, Michael and Pflugfelder, Roman and K{\"a}m{\"a}r{\"a}inen, Joni-Kristian and Chang, Hyung Jin and Danelljan, Martin and Zajc, Luka {\v{C}}ehovin and Luke{\v{z}}i{\v{c}}, Alan and others},
	booktitle=ECCV,
	pages={431--460},
	year={2022},
}

@article{lasher,
	title={LasHeR: A large-scale high-diversity benchmark for RGBT tracking},
	author={Li, Chenglong and Xue, Wanlin and Jia, Yaqing and Qu, Zhichen and Luo, Bin and Tang, Jin and Sun, Dengdi},
	journal=TIP,
	volume={31},
	pages={392--404},
	year={2021},
}

@article{rgbt234,
	title={RGB-T object tracking: Benchmark and baseline},
	author={Li, Chenglong and Liang, Xinyan and Lu, Yijuan and Zhao, Nan and Tang, Jin},
	journal={Pattern Recog.},
	volume={96},
	pages={1856--1864},
	year={2019},
}

@inproceedings{protrack,
	title={Prompting for multi-modal tracking},
	author={Yang, Jinyu and Li, Zhe and Zheng, Feng and Leonardis, Ales and Song, Jingkuan},
	booktitle=ACMMM,
	pages={3492--3500},
	year={2022}
}

\end{document}